\documentclass[letterpaper]{article} % DO NOT CHANGE THIS
\usepackage[draft]{aaai25}  % DO NOT CHANGE THIS
\usepackage{subfigure}
\usepackage{booktabs}
\usepackage{makecell}
\usepackage{color}
\usepackage{times}  % DO NOT CHANGE THIS
\usepackage{helvet}  % DO NOT CHANGE THIS
\usepackage{courier}  % DO NOT CHANGE THIS
\usepackage[hyphens]{url}  % DO NOT CHANGE THIS
\usepackage{graphicx} % DO NOT CHANGE THIS
\usepackage{natbib}  % DO NOT CHANGE THIS AND DO NOT ADD ANY OPTIONS TO IT
\usepackage{caption} % DO NOT CHANGE THIS AND DO NOT ADD ANY OPTIONS TO IT
\usepackage{algorithm}
\usepackage{algorithmic}
\usepackage{threeparttable}
\usepackage{newfloat}
\usepackage{listings}
\DeclareCaptionStyle{ruled}{labelfont=normalfont,labelsep=colon,strut=off} % DO NOT CHANGE THIS
\floatstyle{ruled}
\newfloat{listing}{tb}{lst}{}
\floatname{listing}{Listing}
\title{NeuralMatrix: Compute the Entire Neural Networks with Linear Matrix Operations for Efficient Inference}
\author{
    Ruiqi Sun\textsuperscript{\rm 1},
    Siwei Ye\textsuperscript{\rm 1},
    Jie Zhao\textsuperscript{\rm 2},
    Xin He\textsuperscript{\rm 3},
    Jianzhe Lin\textsuperscript{\rm 2},
    Yiran Li\textsuperscript{\rm 2},
    An Zou\textsuperscript{\rm 1}
    }
\affiliations{
    \textsuperscript{\rm 1}Shanghai Jiao Tong University\\
    \textsuperscript{\rm 2}Microsoft Corporation\\
    \textsuperscript{\rm 3}University of Michigan\\
}

\usepackage{bibentry}
\begin{document}

\maketitle

\begin{abstract}
The inherent diversity of computation types within the deep neural network (DNN) models often requires a variety of specialized units in hardware processors, which limits computational efficiency, increasing both inference latency and power consumption—especially when the hardware processor needs to support and execute different neural networks. In this study, we introduce \textit{NeuralMatrix}, which elastically transforms the computations of entire DNNs into linear matrix operations. This transformation allows seamless execution of various DNN models all with matrix operations and paves the way for running versatile DNN models with a single General Matrix Multiplication (GEMM) accelerator. Extensive experiments with both CNN and transformer-based models demonstrate the potential of \textit{NeuralMatrix} to accurately and efficiently execute a wide range of DNN models, achieving 2.17-38.72× computation efficiency (i.e., throughput per power) compared to CPUs, GPUs, and SoC platforms. This level of efficiency is usually only attainable with the accelerator designed for a specific neural network.
\end{abstract}

\section{Introduction}
\label{sec:introduction}
In recent years, the advancement of deep neural network models has led to their application across a broad spectrum of scenarios. As these neural network architectures grow in size and complexity, they present significant computational challenges, particularly for resource-constrained platforms and budget-conscious organizations.
The Application-Specific Integrated Circuit (ASIC) accelerator offers a promising solution for supporting DNNs on mobile and edge devices. For example, Bai et al. \cite{8438987} introduced a Convolutional Neural Network accelerator design that incorporates a multiplier array, add tree, normalization, ReLU, and pooling units. Similarly, Tambe et al. \cite{edgebert} proposed an edge transformer accelerator featuring processing units (with floating-point vector and accumulate) and dedicated function units for layer normalization, Softmax, and other unique operators in each layer.

As the name (application-specific) indicates, the ASIC accelerators are known for the efficient execution of specific DNN applications. However, their inherent specificity, including the type and number of computation units, can restrict their adaptability from one DNN model to another. For example, transformer-based BERT uses 72.5\% of its computation cycles for versatile nonlinear operations \cite{edgebert}, necessitating the integration of specific types and amounts of nonlinear functional units in its accelerator. However, these functional units can become unnecessary burdens when the same accelerator is used for other networks, such as CNNs, which have far fewer nonlinear operations. A significant gap exists between the generality and computation efficiency of running versatile DNN models on the hardware accelerators \citep{geng2021survey}.

\begin{figure*}
    \centering
    \includegraphics[width = 0.9\textwidth]{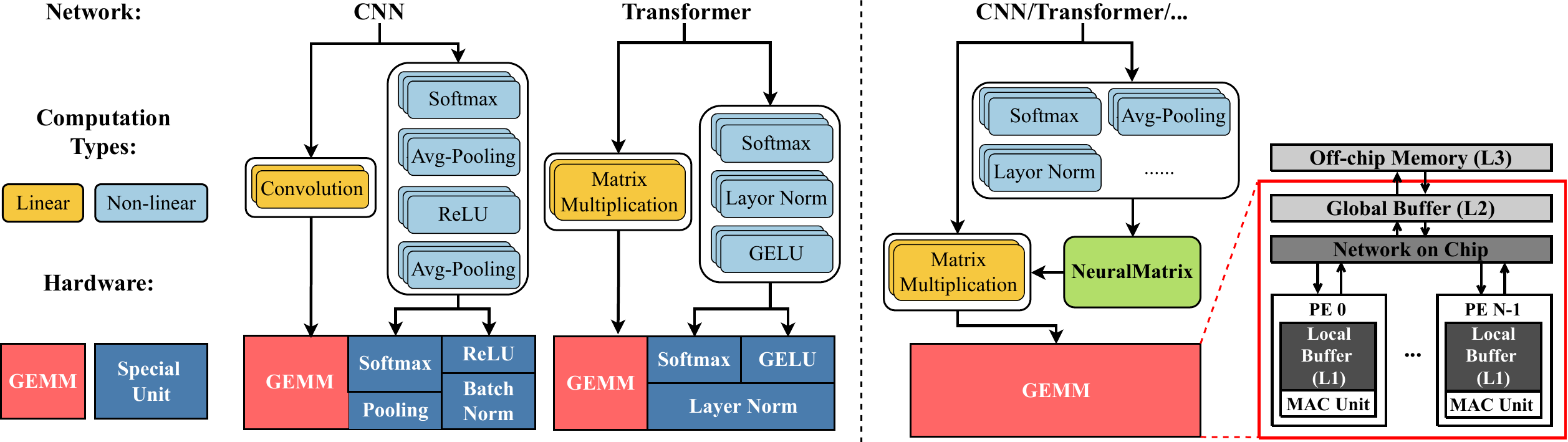}
    \caption{\textit{NeuralMatrix} translates neural network computation into matrix operations, enabling them on a GEMM accelerator.}
    \label{fig:intro}
\end{figure*}

In this study, we present \textit{NeuralMatrix}, a general and compact approach, to efficiently compute the entire neural network with linear matrix operations and seamlessly enable versatile neural networks in a single General Matrix Multiplication (GEMM) accelerator in Fig. \ref{fig:intro}. On one hand, \textit{NeuralMatrix} reduces the computation costs in the DNN models with minimal additional parameters. On the other hand, it overcomes the limitations associated with specific computation types, enabling the generality of running various DNNs on a single GEMM accelerator, which supports general linear matrix operations.
Our comprehensive experiments, utilizing popular backbone models such as CNNs and Transformers, demonstrate that the \textit{NeuralMatrix} enables the generality of running various DNNs with improved computational efficiency (up to 38.72$\times$ throughput per power compared to CPUs, GPUs, and SoC platforms).

% To the best of our knowledge, we are pioneering the transformation of entire diverse DNNs into linear matrix operations, effectively bridging the gap in inference accuracy, computational generality, and efficiency by leveraging a single GEMM accelerator.

%\color{blue}{TODO: cite our date paper: \cite{sun2024onesa}}

\section{Background and Related Work}
\label{sec:background}
\subsection{Intensive / Versatile Computations in DNNs}
The intensive and versatile computations in deep neural networks present significant challenges for computational platforms. Two approaches have been developed to accelerate DNNs.
First, Application-Specific Integrated Circuits (ASICs) are designed for specific network models, achieving optimal efficiency by tailoring functional units to the computations required by the specific model \citep{FPGA_CNN, DaDianNao, li2020ftrans, khan2021npe, wang2021spatten}.
Second, general-purpose AI accelerators, such as GPUs and Tensor Processing Units (TPUs), accelerate DNNs by utilizing numerous processing units capable of handling various types of computations. However, this versatility comes at the cost of higher resource and power consumption \citep{wang2019benchmarking}.

Previous work has attempted to address the complicated and versatility issues stemming from DNNs' nonlinear computations. Approaches such as polynomial fit approximation \citep{DBLP:journals/corr/abs-2101-01321}, Piecewise Linear (PWL) based approximations \citep{dong2020plac, khan2021npe, lyu2021ml, sun2024onesa} and neural network-based approximations \citep{yu2022nn, DBLP:journals/corr/abs-2112-02191} have been proposed to accelerate nonlinear operations in neural networks. Leveraging a neural network to approximate nonlinear operations, an automated approximation framework \citep{lu2023auto} has been developed to simplify and automate this process. MA-BERT \citep{ming2022ma} replaces complex functions with computation-friendly ones in the transformer-based BERT network, using matrix arithmetic and trivial ReLU operations. Experiments show that this substitution significantly can improve computational efficiency on CPUs and GPUs but still hard to avoid additional function units in the accelerator to execute the newly introduced computation.

\subsection{General Matrix Multiplication Accelerator}
The General Matrix Multiplication (GEMM) accelerator is specialized hardware circuit designed to process matrix multiplication operations \citep{kwon2019understanding}, and its basic architecture is shown in Fig.\ref{Algorithm:Overall framwork}. They are employed in data centers for high-performance computing and edge devices to enhance efficiency in tasks such as digital signal processing, artificial intelligence, and scientific simulations \citep{qin2020sigma}. GEMM accelerators can be integrated with devices like TPU \citep{jouppi2017datacenter}, included in SoC configurations \citep{mitra2014implementation}, or developed as standalone chips \citep{reggiani2023mix}.
Compared to the general-purpose processor CPUs and GPUs, which accommodate a variety of instructions through a range of logic and arithmetic components, GEMM accelerators are explicitly designed for matrix multiplication using only Multiply-Accumulate (MAC) units and buffers. This focused approach to matrix multiplication results in exceptional efficiency \citep{hojabr2021spaghetti}. However, the GEMM accelerator can only process the general matrix multiplication computation. Additional special function units have to be located alongside the GEMM accelerator to process the other types of computations \citep{jouppi2017datacenter, mitra2014implementation}. The types and numbers of special function units are carefully tailored to the computations of the targeted neural network models \cite{pati2021demystifying}. By modifying the hardware architecture of the GEMM accelerator to support piecewise linear approximation, \citep{sun2024onesa} can enable GEMM to handle nonlinear operations in DNN models. However, the design focuses only on hardware implementation, neglecting efficient mapping of DNN operations to the GEMM accelerator, leading to obvious accuracy loss and increased memory consumption. This work proposes \textit{NeuralMatrix}, which elastically transforms the computations of entire DNNs into linear matrix operations and enables accurate and efficient execution on a GEMM accelerator.

\begin{figure*}[t!]
    \centering
    \includegraphics[width = 0.72\textwidth]{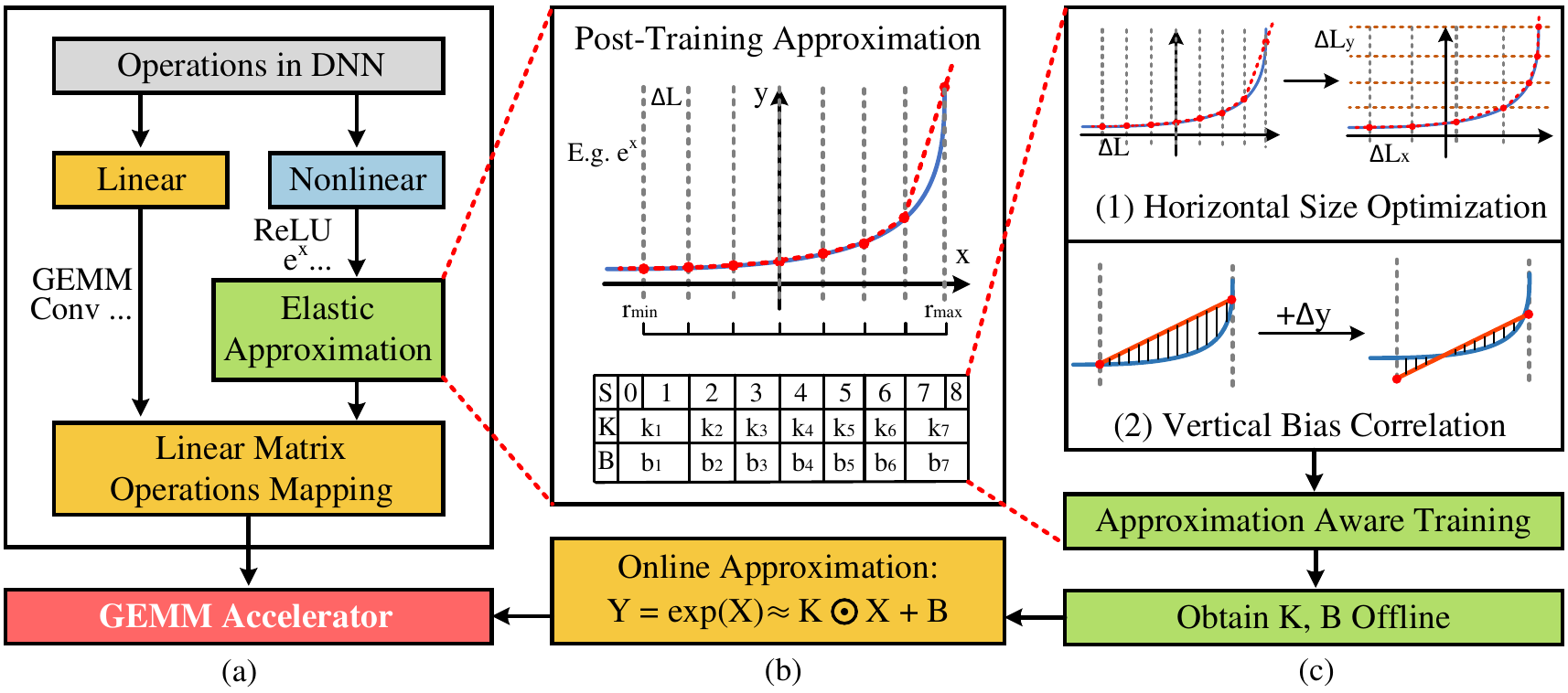}
    \caption{Overview of \textit{NeuralMatrix}: Different types of computation in DNN will go through different decision and process steps. An entire neural network can be moved to linear matrix operations and become fully executed by a GEMM accelerator.}
    \label{fig:overview}
\end{figure*}

% It has been demonstrated that optimizing computation types can improve efficiency for individual networks; however, designs using polynomial fitting and neural network approximation often introduce substantial computational burdens. In contrast, this work proposes a general approach, NeuralMatrix, which elastically transforms various DNN operations into linear matrix operations, and facilitates efficient execution through GEMM accelerators without the need for specialized function units. This results in faster inference times and decreased power consumption. 
% The framework consists five steps: \textbf{1}. Post-Training-Approximation, \textbf{2}. Expectation Bias Correction, \textbf{3}. Parameter Size Optimization, \textbf{4}. Segment Length Optimization and \textbf{5}.

% to compute the versatile computations all with linear matrix operations. The NeuralMatrix enables running versatile neural networks on one GEMM accelerator by eliminating the limitations of deploying the special function units for various computation types. This approach endows different neural network models with application-specific efficiency, which is conventionally only available with application-specific accelerators.
\section{NeuralMatrix -- Computing Networks with Matrix Operations}
\label{sec:mapping}

This section describes how \textit{NeuralMatrix} maps and computes the neural networks with linear matrix operations, which is depicted by the flowchart in Fig.~\ref{fig:overview}. 
First, the computation in neural networks is classified into linear and nonlinear operations. Linear operations are directly mapped to General Matrix Multiplication (GEMM)  accelerators through GEMM mapping (\S~\ref{sec:method-linear}). 
For nonlinear operations, \textit{NeuralMatrix} will then decide if one operation already corresponds to a Piecewise Linear (PWL) function (e.g., ReLU), which can be computed using the PWL calculation method. 
If not, an elastic approximation, with horizontal size optimization and vertical bias correction, will be performed to convert the nonlinear operations to linear matrix operation (\S~\ref{sec:method-nonlinear}). 
To preserve network accuracy after approximation and fine-tuning, we introduce approximation-awared training for \textit{NeuralMatrix} (\S~\ref{sec:train}).

\subsection{Mapping Linear Operations to General Matrix Operation}
\label{sec:method-linear}
Linear operations are pervasive in DNNs, for example in fully connected layers, convolution kernels, attention mechanisms, and more. These linear operations involve 2D, 3D, or higher-dimensional tensors.  
By applying reshaping and re-blocking techniques, these linear operations can be represented as matrix addition and multiplication operations with various sizes. The dimensions of these matrices are determined by the width, height, and number of channels in original convolution computation.

%\begin{algorithm}[t]
%    \renewcommand{\algorithmicrequire}{\textbf{Input:}}
%    \renewcommand{\algorithmicensure}{\textbf{Output:}}
%    \caption{Elastic Approximation}
%    \label{Algorithm:Overall framwork}
%    \begin{algorithmic}
%        \REQUIRE
%            Nonlinear function operation:$f(x)$;
%            Segment length in x:$\Delta L_{x}$;
%            Segment length in y:$\Delta L_{y}$;
%            Expectation threshold:$E_{th}$;
%            Accuracy loss threshold:$Acc_{th}$.
%        \ENSURE
%            Parameter set $K = \{k_{1}, k_{2}, ... k_{n}\}$ and $B = \{b_{1}, b_{2}, ... b_{n}\}$;
%        \STATE // Obtain the range of $f(x)$.
%        \STATE $[r_{min}, r_{max}] \leftarrow$Inference DNN model.
%        \STATE // Obtain parameters k and b in each segment.
%        \STATE for $i=1$ to $n$ :
%        \STATE \qquad $k_{i}, b_{i} \leftarrow$ \textbf{Vertical Bias Correction}
%        \STATE // Obtain the start points set.
%        \STATE $X \leftarrow$ \textbf{Horizontal Size Optimization}
%        \STATE $Acc\_loss \leftarrow$ Inference DNN model with approximation.
%        \STATE if $Acc\_loss > Acc_{th}$ :
%        \STATE \qquad New model $\leftarrow$\textbf{Approximation Aware Training}
%        \STATE \textbf{Return} K and B
%    \end{algorithmic}
%\end{algorithm}

Given that each GEMM accelerator has its own computational and memory capabilities, matrices of different sizes—reshaped from linear operations in DNNs—are processed block-wise on the GEMM accelerator. In other words, the input and weight matrices are partitioned into smaller blocks to compute the output matrix, taking advantage of the GEMM accelerator's three-level memory hierarchy to minimize energy consumption and buffer access times \citep{kwon2019understanding}.
% \jie{the following paragraph is a little confusing to me} The optimal block division is achieved by exploring data flows following a top-down approach: the stationary scheme, then spatial/temporal accesses, and finally tile size to find the optimized data flow of linear operations. The term "stationary" refers to storing the matrix data in global and local buffers for the longest time to maximize its reuse. Data reuse can then be classified into temporal and spatial reuse. Temporal reuse involves reading the data from off-chip DRAM in chronological order, sending it to multiple local buffers, and multiplying or adding up the sum of partial sums in processing elements (PEs). On the other hand, spatial reuse means moving and calculating data in parallel. Finally, the tile size defines the data size in each movement and computation. Given a GEMM accelerator, exploring the block division optimally, similar to a grid search, can find the efficient mapping of network linear operations to the GEMM accelerators.
The optimal block division is achieved by exploring data flows using a top-down approach: first addressing the stationary scheme, followed by spatial/temporal accesses, and finally determining the tile size to find the optimized data flow of linear operations. The term ``stationary'' refers to storing matrix data in global and local buffers for extended periods to maximize its reuse. Data reuse can be classified into temporal and spatial reuse.
Temporal reuse involves reading data from off-chip Dynamic Random Access Memory (DRAM) in chronological order, sending it to multiple local buffers, and performing multiplication or addition operations on the partial sums in Processing Elements (PEs). Conversely, spatial reuse entails moving and processing data in parallel. Lastly, the tile size defines the data size for each movement and computation. 

The above division uses a method similar to grid search to find this optimal block division. For example, given a matrix multiplication with dimension $(M \times K)\times(K \times N)$, we change the block size in the three dimensions (stationary, spatial/temporal accesses, and tile sizes) from 2 to 128 with stride 2, and use an early stage model to calculate the latency and energy consumption of GEMM accelerator. Then we will choose the optimal block size in three dimensions with the minimum latency or energy.

\subsection{Elastic Approximation for Nonlinear Operations}
\label{sec:method-nonlinear}
Addressing the nonlinear operations inherent in DNNs poses a significant challenge, as they cannot be easily mapped to standard matrix operations. To overcome this issue, the  \textit{NeuralMatrix} introduce elastic approximation to strategically moves nonlinear operations to matrix operations. This approach ensures the dual benefits of accuracy and computation efficiency, achieved through a sequence of steps: initial post-training approximation, followed by horizontal size optimization and vertical bias correction, which is summarized in Alg. \ref{Algorithm:Overall framwork}.

\subsubsection{Post-Training Approximation}
In the proposed elastic approximation, complex nonlinear operations are first approximated using a continuous PWL function, which can be shown in Fig. \ref{fig:overview}(b):
\begin{itemize}
    \item[1)]
    We first obtain the input range of this non-linear operation by running the DNN model.
    \item[2)]
    Then divide the non-linear function into several small segments $(s_{1}, s_{2} ... s_{7})$ with fixed length, and fit the original function in each segment with a linear operation: $y=exp(x) \approx k_{i}*x + b_{i}$, ensuring that the endpoint of one segment is the same as the starting point of the next segment. We use another two segments $s_{0}, s_{8}$ to represent the input outside the range and they share the same parameter k, b with $s_{1}$ and $s_{7}$ respectively.
\end{itemize}

\renewcommand{\algorithmicrequire}{\textbf{Input:}}
\renewcommand{\algorithmicensure}{\textbf{Output:}}
\begin{algorithm}[t]
\begin{small}
    \begin{algorithmic}
        \REQUIRE
            Nonlinear function operation:$f(x)$;
            Segment length in x:$\Delta L_{x}$;
            Segment length in y:$\Delta L_{y}$;
            Expectation threshold:$E_{th}$;
            Accuracy loss threshold:$Acc_{th}$.
        \ENSURE
            Parameter set $K = \{k_{1}, k_{2}, ... k_{n}\}$ and $B = \{b_{1}, b_{2}, ... b_{n}\}$;
        \STATE // Obtain the range of $f(x)$.
        \STATE $[r_{min}, r_{max}] \leftarrow$Inference DNN model.
        \STATE // Obtain and optimize the start points set.
        \STATE $X \leftarrow$ \textbf{Horizontal Size Optimization}
        \STATE // Obtain parameters k and b in each segment.
        \STATE for $i=1$ to $n$ :
        \STATE \qquad $k_{i}, b_{i} \leftarrow$ \textbf{Vertical Bias Correction}
        \STATE $Acc\_loss \leftarrow$ Inference DNN model with approximation.
        \STATE if $Acc\_loss > Acc_{th}$ :
        \STATE \qquad New model $\leftarrow$\textbf{Approximation Aware Training}
        \STATE \textbf{Return} K and B
    \end{algorithmic}
\end{small}
\caption{Elastic Approximation}
\label{Algorithm:Overall framwork}
\end{algorithm}

\begin{figure}
     \centering  
        \begin{minipage}[c]{0.235\textwidth}
            \captionsetup{type=subfigure}
            \includegraphics[width=\textwidth]{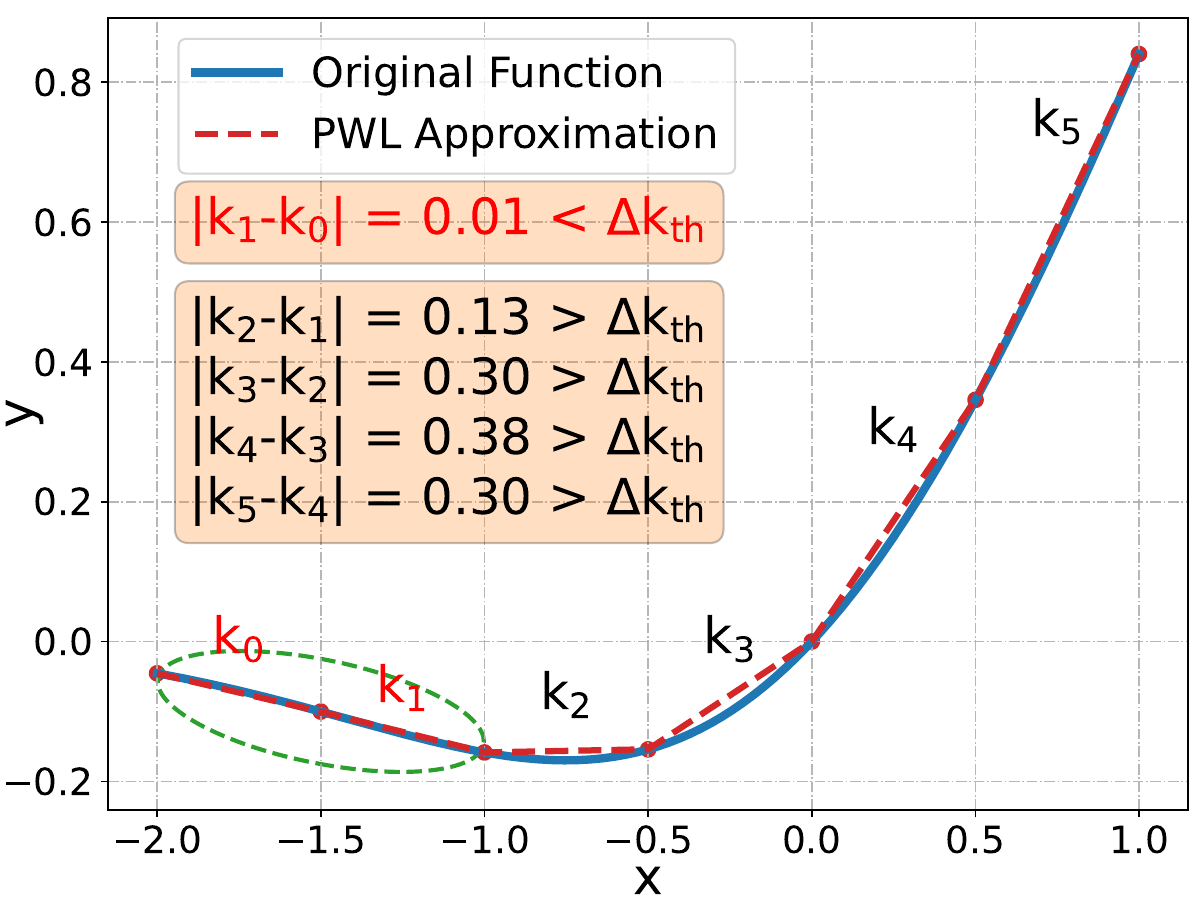}
            \caption{Size Optimization}
        \end{minipage}
        \hspace{-1.5mm}
        \begin{minipage}[c]{0.235\textwidth}  
            \captionsetup{type=subfigure}
            \includegraphics[width=\textwidth]{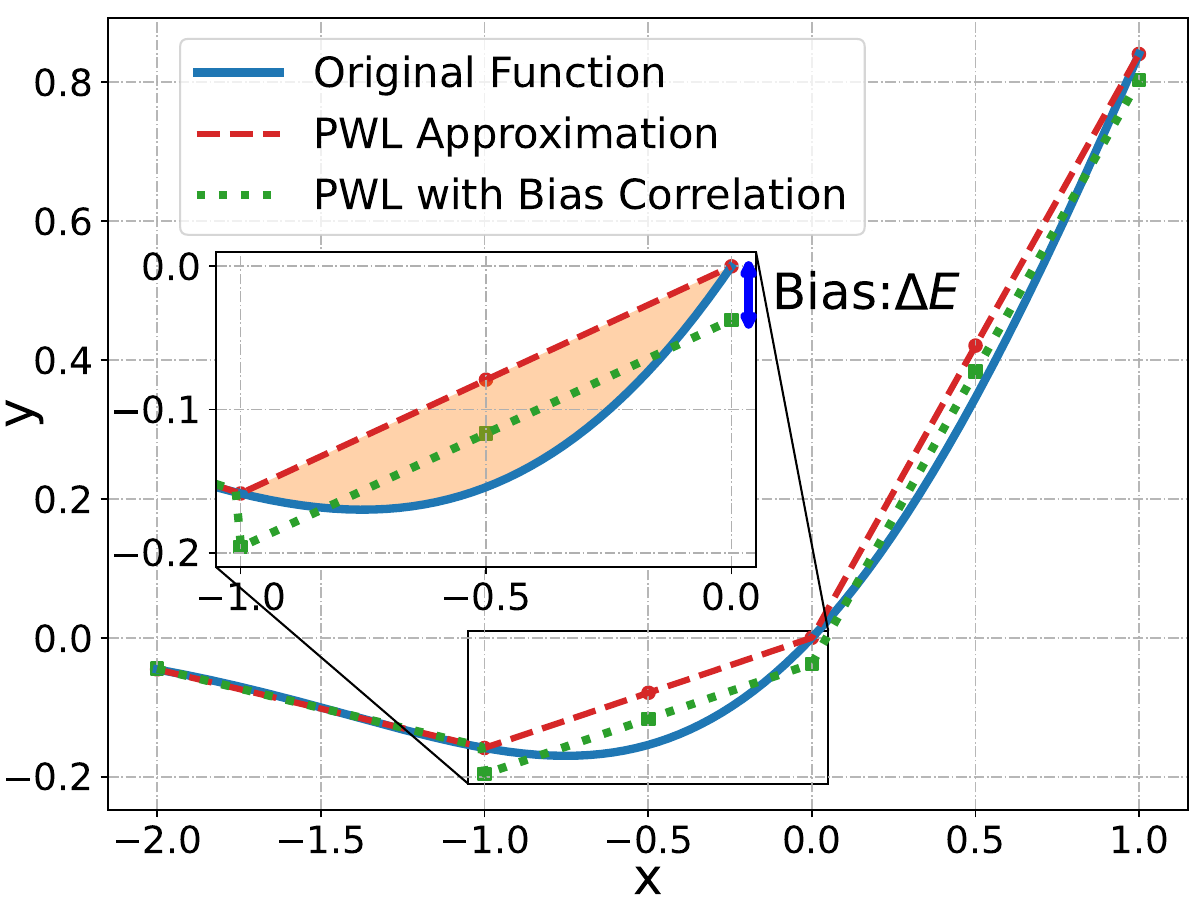}
            \caption{Bias Correction}
        \end{minipage}
        
    \caption{A showcase to the process of Elastic Approximation for activation function GELU.}
    \label{fig:Mapping}
\end{figure}

\renewcommand{\algorithmicrequire}{\textbf{Input:}}
\renewcommand{\algorithmicensure}{\textbf{Output:}}
\begin{algorithm}[t]
\begin{small}
    \begin{algorithmic}
        \REQUIRE
            Nonlinear function operation:$f(x)$;
            Primary segment length in x: $\Delta L_{x}$;
            Threshold for slope change: $\Delta K_{th}$;
            Range of nonlinear function: $[r_{min}, r_{max}]$.
        \ENSURE
            Start points set $X = \{x_{1},x_{2}, ... x_{n}\}$.
        \STATE $i = 1$, $x_{i} = r_{min} + \Delta L_{x}, k=k_0=\frac{f(x_i) - f(r_{min})}{\Delta L_{x}}$
        \STATE while($x_{i} < r_{max}$) :
        \STATE \qquad $k_{i} = \frac{f(x_i+\Delta L_{x}) - f(x_{i})}{\Delta L_{x}}$
        \STATE \qquad if $\vert k_i -k\vert \leq \Delta K_{th}$ :
        \STATE \qquad // Two segments can be merged into one.
        \STATE \qquad\qquad Pass
        \STATE \qquad else:
        \STATE \qquad\qquad Add $x_{i}$ to X,  $k = k_i, i++$
        \STATE \qquad $x_{i} = x_{i} + \Delta L_{x}$
        \STATE \textbf{Return} $X = \{r_{min},x_{1},x_{2}, ... x_{n}\}$
    \end{algorithmic}
\end{small}
\caption{Horizontal Size Optimization}
\label{Algorithm:Parameter Size Optimization}
\end{algorithm}

%The pre-calculated parameters $k$ and $b$ of each segment are pre-stored in the processor memory (i.e., global buffer of the GEMM accelerator), and indexed by the segment numbers. \textbf{\scriptsize \Circled{3}} To calculate the piecewise linear operations in accelerator, we will use a linear operator to calculate the segment matrix $S$ for the input matrix $X$. Each element in $S$, e.g., $S_{i,j}$ represents which segment its corresponding input value $X_{i,j}$ falls into. Then parameters $k$ and $b$ are aggregated and sent to the GEMM accelerator in the forms of slope matrix $K$ and intercept matrix $B$. Finally, the GEMM accelerator performs element-wise calculations $Y = X \cdot K+B$ to get the output matrix $Y$.

%There are two primary advantages of employing CPWL in NeuralMatrix. Firstly, classic GEMM accelerators can naturally support the computation in CPWL, unlike some other approximation methods, such as look-up table (LUT) based nonlinear approximation, which require extra hardware resources. Secondly, alternative approximation methods like Taylor expansion or neural network approximation necessitate considerable additional computations and parameters, which do not align with our ultimate goal of computing efficiency.(\textcolor{blue}{ADD AN EXPERIMENT TO COMPARE THE COMPUTATION AND PARAMETER OF DIFFERENT APPROXIMATION METHODS})

\subsubsection{Horizontal Size Optimization} In the above Post-Training Approximation, the length of every segment is set as a fixed value, but this is not optimized for all kinds of nonlinear operations. For example, the activation function Gaussian Error Linear Unit (GELU) is close to line $y=0$ and $y=x$ when $x\leq-3$ and $x\geq3$ respectively, so we can use fewer parameters to approximate the original function in these two regions and more parameters to approximate in the region $-3 \leq x \leq3$. 
However, if we set the segment length as a fixed value, it may lead to unnecessary consumption of memory or buffer sizes to store the parameters in $x\leq-3$ and $x\geq3$. Therefore to better assign the length of each segment, we propose a horizontal size optimization algorithm, and the detailed description is shown in Alg. \ref{Algorithm:Parameter Size Optimization}. In this algorithm, we first assign two hyperparameters $\Delta L_{x}$ and $\Delta K_{th}$, and if the function changes slightly, the difference in slope between the two segments is less than the threshold, we can combine these two segments into one. If the function changes rapidly, the slope difference between two segments will exceed the threshold, then the starting point of this segment will be added to the start point set for the PWL approximation. Fig.\ref{fig:Mapping}(a) shows an example of this optimization process. In this case, we set the threshold for slope change to 0.1 and primary segment length as 0.5, and due to the original function changes slightly in the region $-2 \leq x \leq -1$, these two segments can be merged into one which reduces the parameters used for approximation.

The advantages of this method are obvious, an optimized segment length assignment not only improves approximation performance but also reduces the number of segments, meaning fewer model parameters. We applied our method to four common nonlinear functions in DNN model: exponential($e^x$), square root($\sqrt{x}$), Inverse proportional($\frac{1}{x}$), and activation function GELU, and the result shows that the parameter consumption can be reduced by 37.88\%, 51.56\%, 66.75\%, and 63.50\% without mean square error (MSE) decreases.

\begin{algorithm}[t]
\begin{small}
    \begin{algorithmic}
        \REQUIRE
            Nonlinear function operation:$f(x)$;
            Start point of segment:$x_{i}$;
            End point of segment:$x_{i+1}$;
            Expectation threshold:$E_{th}$.
        \ENSURE
            Parameters: $k_{i}$ and $b_{i}$
        \STATE // Calculate the parameter for PWL approximation.
        \STATE $k_{i} = \frac{f(x_{i+1}) - f(x_{i})}{x_{i+1} - x_{i} }$, $b_{i} = f(x_{i})-k_{i}*x_{i}$
        \STATE // Calculate the expectation bias caused by PWL approximation.
        \STATE $\Delta E = \frac{1}{x_{i+1} - x_{i}} \int_{x_{i}}^{x_{i+1}} {f(x)} \,{\rm d}x - \frac{1}{2} (f(x_{i}) + f(x_{i+1}))$
        \STATE // Expectation bias correlation.
        \STATE if $\Delta E \leq E_{th}$ : $b_{i} = b_{i} + \Delta E$
        \STATE \textbf{Return} $k_{i}, b_{i}$
    \end{algorithmic}
\end{small}
\caption{Vertical Bias Correlation}
\label{Algorithm:Expectation Bias Correction}
\end{algorithm}

\subsubsection{Vertical Bias Correction}
Although PWL approximation can significantly reduce the computation brought by nonlinear operations, it suffers from a low approximation accuracy especially for those rapidly changing functions (e.g. $e^{x}, \frac{1}{x}$). For example in Fig. \ref{fig:overview}(b), the PWL approximation of $e^{x}$ deviates critically from the original function in $7^{th}$ segment, which may lead to a serious accuracy drop to the DNN model.

\begin{figure*}[t]
\begin{center}
\setlength{\abovecaptionskip}{-0.1cm}
\hspace{4mm}
\includegraphics[trim=0cm 0cm 0cm 0cm, clip,width=0.9\textwidth]{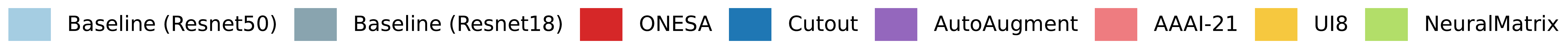}

\subfigure[CIFAR10]{
    %\label{fig:linear} %% label for first subfigure
    \includegraphics[trim=0cm 0cm 0cm 0cm, clip,width=0.23\textwidth]{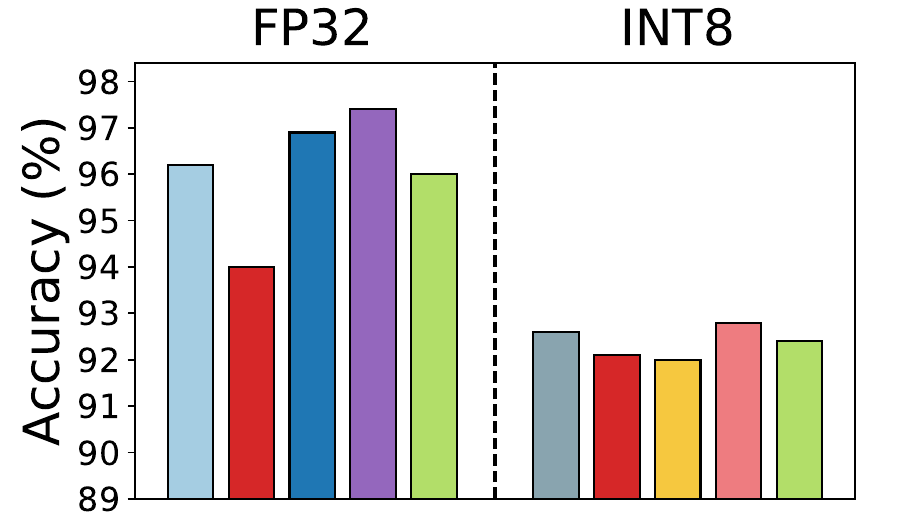}
}
\hspace{-4mm}
\subfigure[CIFAR100]{
    %\label{fig:nonlinear} %% label for second subfigure 
    \includegraphics[trim=0cm 0cm 0cm 0cm, clip,width=0.23\textwidth]{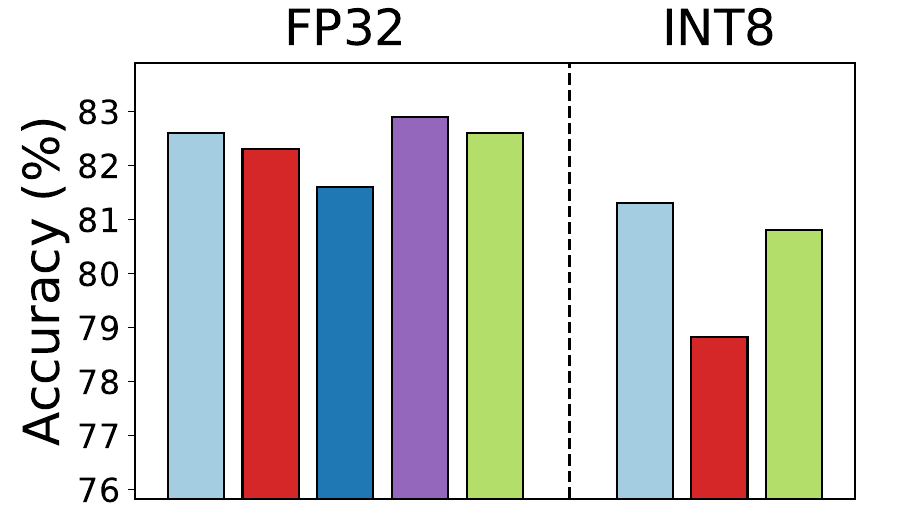}
}
\hspace{-4mm}
\subfigure[ImageNet]{
    %\label{fig:nonlinear} %% label for second subfigure 
    \includegraphics[trim=0cm 0cm 0cm 0cm, clip,width=0.23\textwidth]{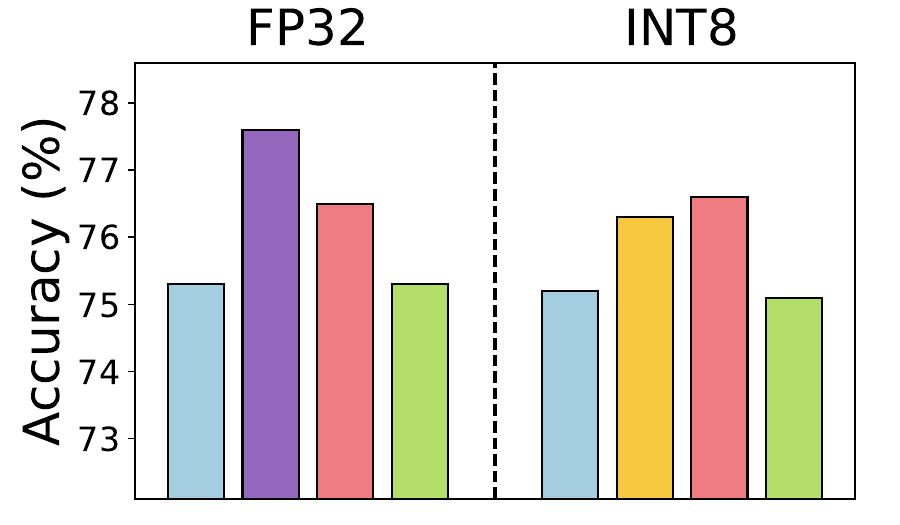}
}
\hspace{-4mm}
\subfigure[MNIST-Fashion]{
    %\label{fig:nonlinear} %% label for second subfigure 
    \includegraphics[trim=0cm 0cm 0cm 0cm, clip,width=0.23\textwidth]{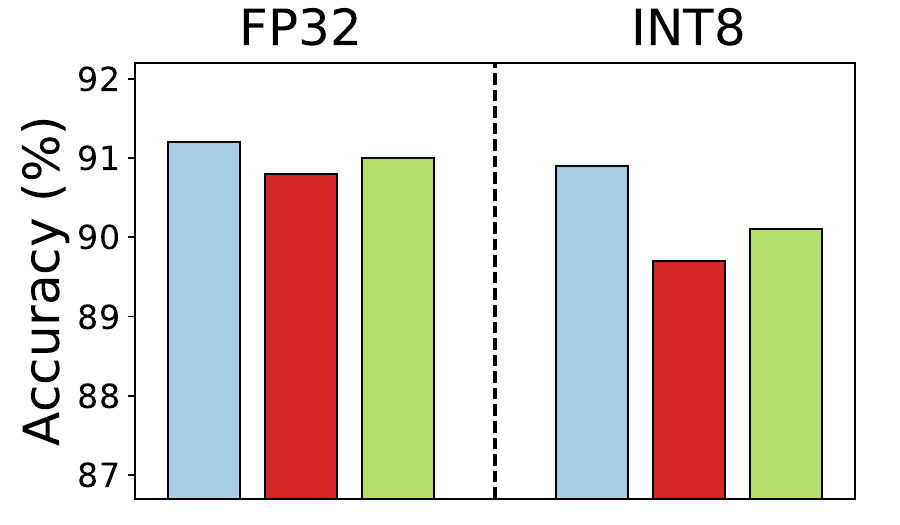}
}
\caption{Network accuracy with CNN-based ResNet.}
\label{fig:accuracy_cnn}
\end{center}
\end{figure*}

\begin{figure*}[t]
\begin{center}
\setlength{\abovecaptionskip}{-0.1cm}
\includegraphics[trim=0cm 0cm 0cm 0cm, clip,width=0.9\textwidth]{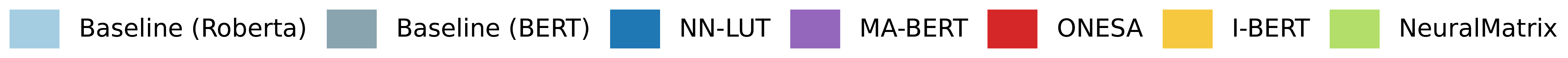}
\hspace{-4mm}
\subfigure[CoLA]{
    %\label{fig:linear} %% label for first subfigure
    \includegraphics[trim=0cm 0cm 0cm 0cm, clip,width=0.23\textwidth]{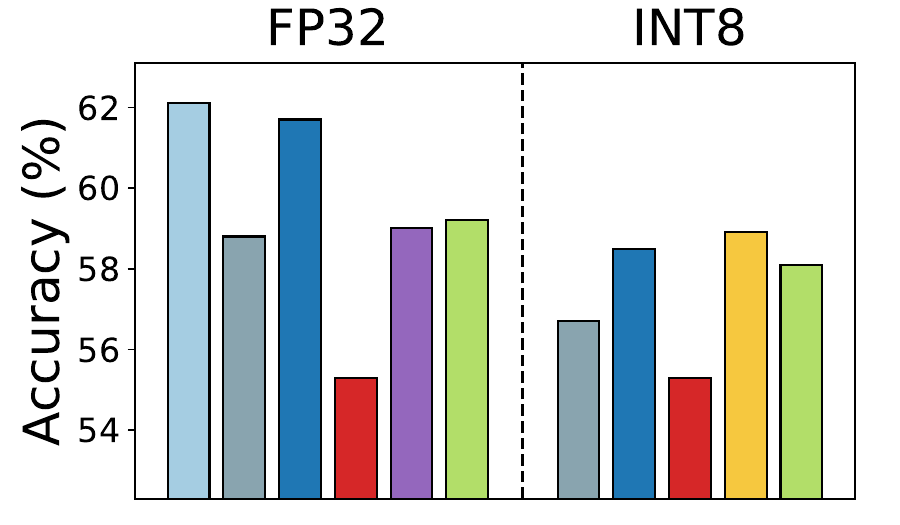}
}
\hspace{-4mm}
\subfigure[MNLI]{
    %\label{fig:nonlinear} %% label for second subfigure 
    \includegraphics[trim=0cm 0cm 0cm 0cm, clip,width=0.23\textwidth]{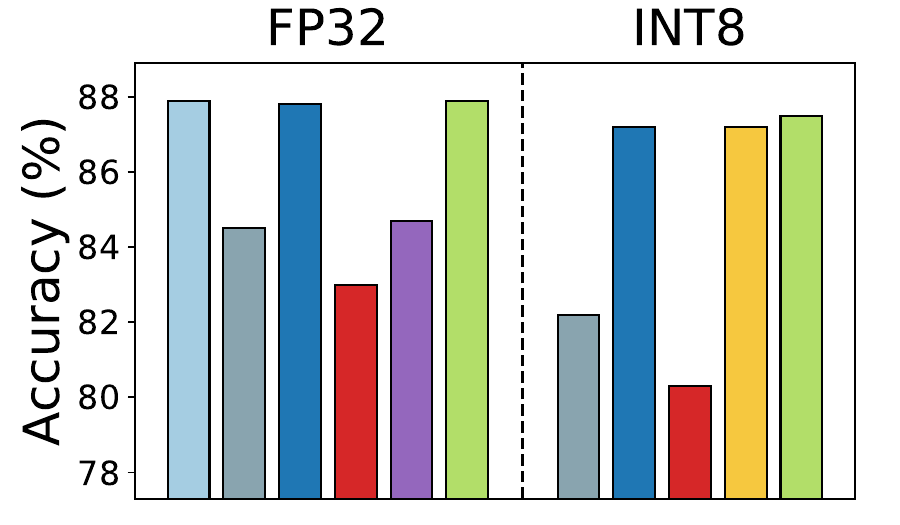}
}
\hspace{-4mm}
\subfigure[MRPC]{
    %\label{fig:nonlinear} %% label for second subfigure 
    \includegraphics[trim=0cm 0cm 0cm 0cm, clip,width=0.23\textwidth]{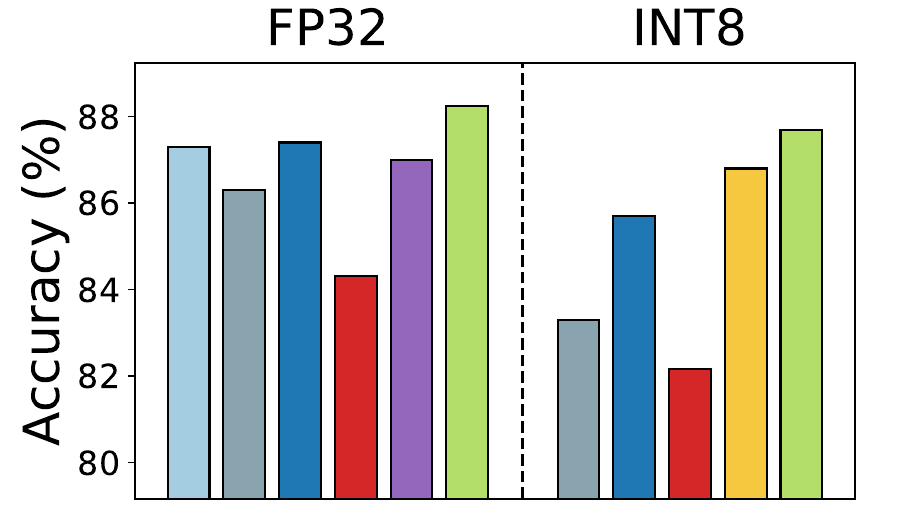}
}
\hspace{-4mm}
\subfigure[QNLI]{
    %\label{fig:nonlinear} %% label for second subfigure 
    \includegraphics[trim=0cm 0cm 0cm 0cm, clip,width=0.23\textwidth]{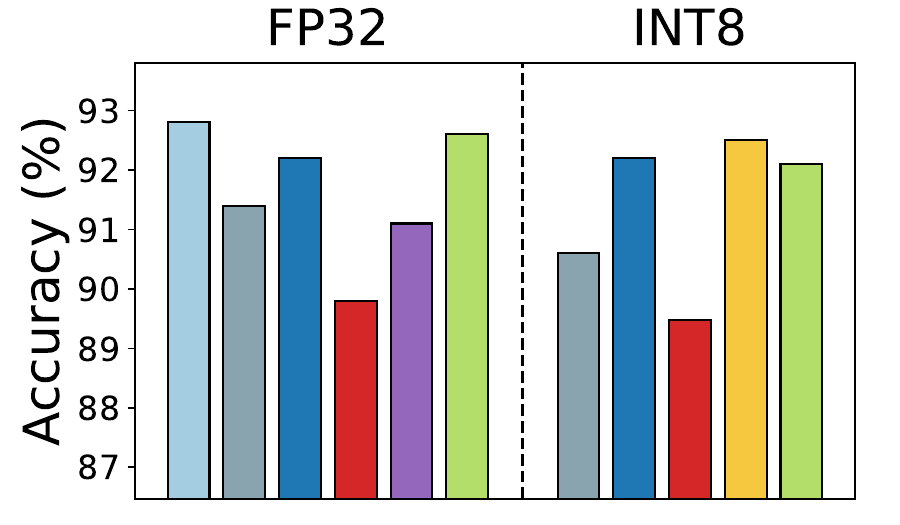}
}

\subfigure[QQP]{
    %\label{fig:nonlinear} %% label for second subfigure 
    \includegraphics[trim=0cm 0cm 0cm 0cm, clip,width=0.23\textwidth]{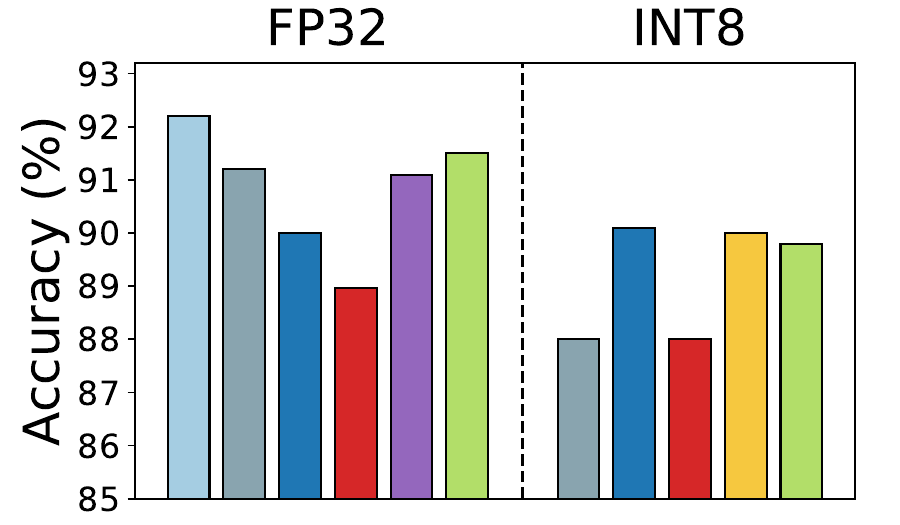}
}
\hspace{-4mm}
\subfigure[RTE]{
    %\label{fig:nonlinear} %% label for second subfigure 
    \includegraphics[trim=0cm 0cm 0cm 0cm, clip,width=0.23\textwidth]{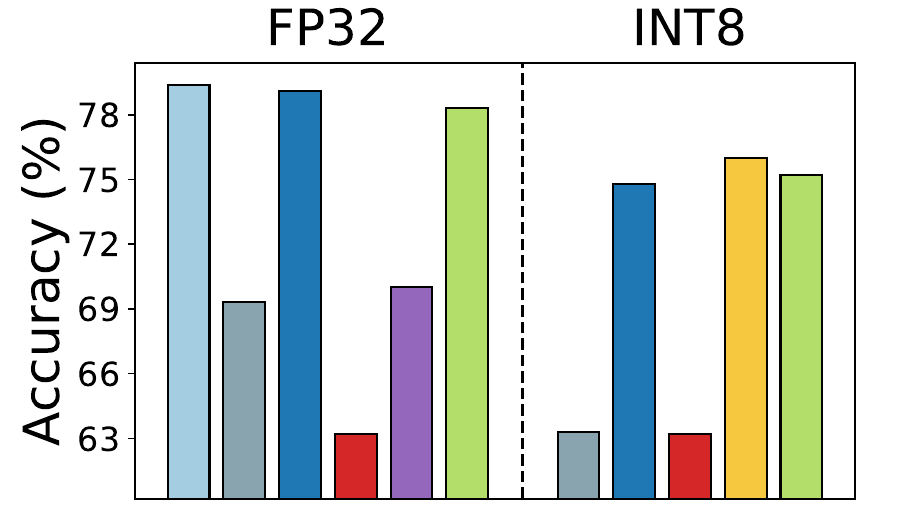}
}
\hspace{-4mm}
\subfigure[STS-2]{
    %\label{fig:nonlinear} %% label for second subfigure 
    \includegraphics[trim=0cm 0cm 0cm 0cm, clip,width=0.23\textwidth]{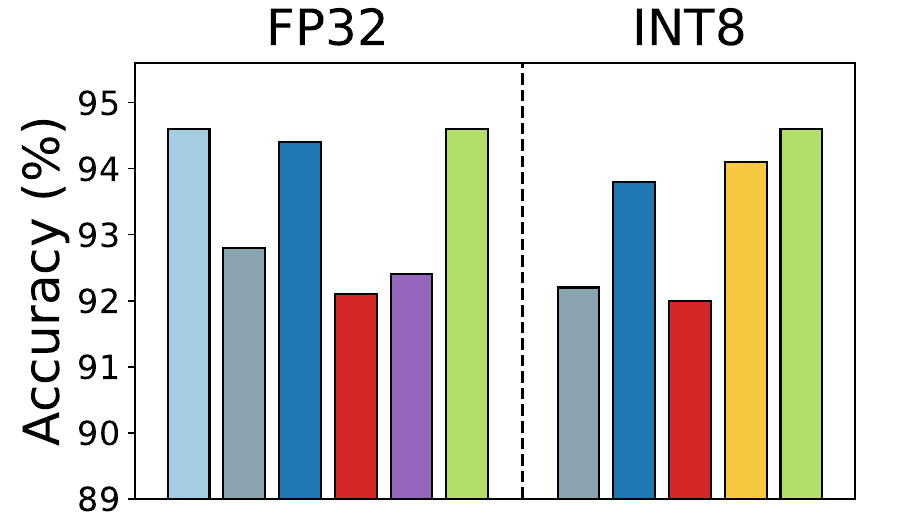}
}
\hspace{-4mm}
\subfigure[STS-B]{
    %\label{fig:nonlinear} %% label for second subfigure 
    \includegraphics[trim=0cm 0cm 0cm 0cm, clip,width=0.23\textwidth]{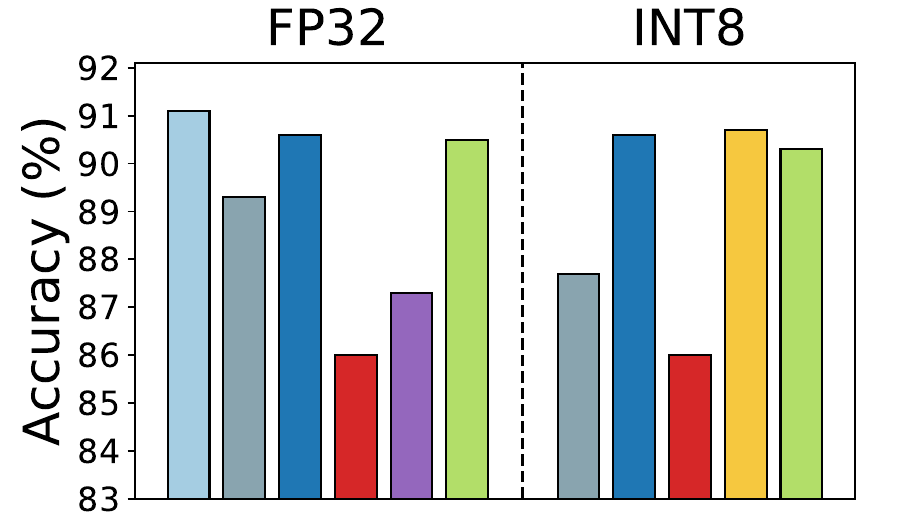}
}

\caption{Network accuracy with transformer-based BERT and RoBERTa.}
\label{fig:accuracy_bert}
\end{center}
\end{figure*}

\renewcommand{\algorithmicrequire}{\textbf{Input:}}
\renewcommand{\algorithmicensure}{\textbf{Output:}}

To enhance the approximation, we propose a vertical bias correlation method, and the detailed algorithm is described in Alg. \ref{Algorithm:Expectation Bias Correction}. In this method, we calculate the expectation of the approximated function and original function in each segment by integration, and if their difference $\Delta y$ exceeds the threshold we will add the difference to parameter $b_{i}$. This process can be shown in Fig.\ref{fig:Mapping} (b), and it's obvious that the conventional PWL approximation will lead to a severe expectation bias in segments ranging from -1 to 0. Our bias correlation can address this issue by offsetting the conventional method with $\Delta E$, and this value can be calculated by dividing the shaded area between the curve obtained from the PWL approximation and the original function by the segment length. The whole process of bias correlation can be completed offline, so it won't lead to additional computation and parameter consumption to the online DNN inference. Our experiment shows that, under the condition of using the same parameters to approximate nonlinear functions within the same input range, bias correlation can reduce MSE of approximation to exponential($e^x$), square root($\sqrt{x}$), Inverse proportional($\frac{1}{x}$) and activation function GELU by 81.67\%, 83.18\%, 73.97\% and 82.28\% respectively.

\section{Approximation-Aware Training}
\label{sec:train}
Following the previous section, the DNN is transformed into linear matrix operations with elastic approximation. After this transformation, the \textit{NeuralMatrix} can be seamlessly integrated with training, offering an approximation-aware training approach. This technique involves mapping a pre-trained DNN to its approximated form and then fine-tuning it on specific downstream tasks. The loss function used during approximation-aware training remains unchanged from conventional fine-tuning, and standard automatic differentiation techniques can be employed for back-propagation.

% Any approximation method that maps DNNs, including nonlinear functions, to the GEMM accelerator will inevitably lead to a loss in computational accuracy, which in turn results in end-to-end inference accuracy loss. In the approximation of nonlinear operations, the accuracy of the approximation is dependent on the granularity of linear segments. Finer granularity contributes to higher accuracy but increases the number of parameters stored in the memory. In Sec.~\ref{sec:eval-parameter}, we will demonstrate how to find a suitable tradeoff and select the parameter granularity to achieve both low memory cost and high inference accuracy for various downstream tasks.

% Based on the experimental results presented in Section~\ref{sect:eval-accuracy}, which compare the inference accuracy of different benchmarks using various DNN architectures, we advocate for the pre-approximation method over the post-approximation approach.
\section{Model Performance and Computation Cost}
\label{sect:eval}

This section verifies the inference accuracy and model parameter sizes after different DNNs are transformed into linear matrix operations using the proposed \textit{NeuralMatrix}. Next, we compare the computational efficiency with a FPGA-based General Matrix Multiplication (GEMM) accelerator with existing general-purpose and application-specific computation platforms. This evaluation proves the benefits of \textit{NeuralMatrix} in terms of accuracy, model size, and computational efficiency.

We experiment with two main DNN categories: CNN-based ResNet \cite{he2016deep} and transformer-based BERT \cite{devlin2018bert} and RoBERTa \cite{liu2019roberta}, with both Floating-Point (FP32) and Integer (INT8) quantization of each network model. For CNN-based ResNet, we use CIFAR-10 \cite{krizhevsky2009learning}, CIFAR-100 \cite{krizhevsky2009learning}, Mnist-Fashion \cite{xiao2017fashion}, and Imagenet \cite{deng2009imagenet} datasets. For transformer-based BERT, we use the General Language Understanding Evaluation (GLUE) benchmark \cite{wang2018glue}, which consists of language understanding tasks at difficulty levels.

\subsection{Inference Accuracy}
\label{sec:eval-accuracy}
We first verify the inference accuracy across two popular DNN architecture categories. Fig. \ref{fig:accuracy_cnn} and \ref{fig:accuracy_bert} present the network inference accuracy for the transformer-based BERT (and its variants) and the CNN-based ResNet, respectively. 

\begin{figure}
     \centering  
        \begin{minipage}[c]{0.235\textwidth}
            \captionsetup{type=subfigure}
            \includegraphics[width=\textwidth]{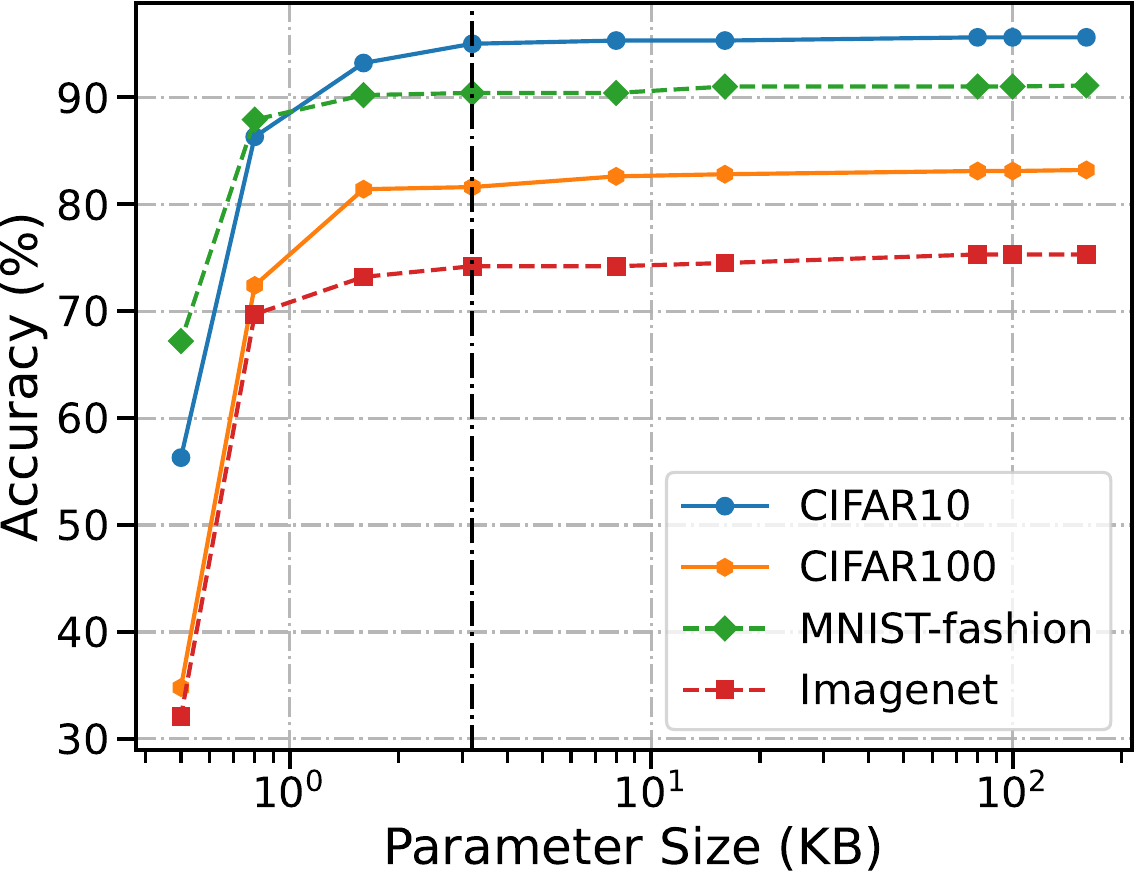}
            \caption{ResNet50}
        \end{minipage}
        \hspace{-1.5mm}
        \begin{minipage}[c]{0.235\textwidth}  
            \captionsetup{type=subfigure}
            \includegraphics[width=\textwidth]{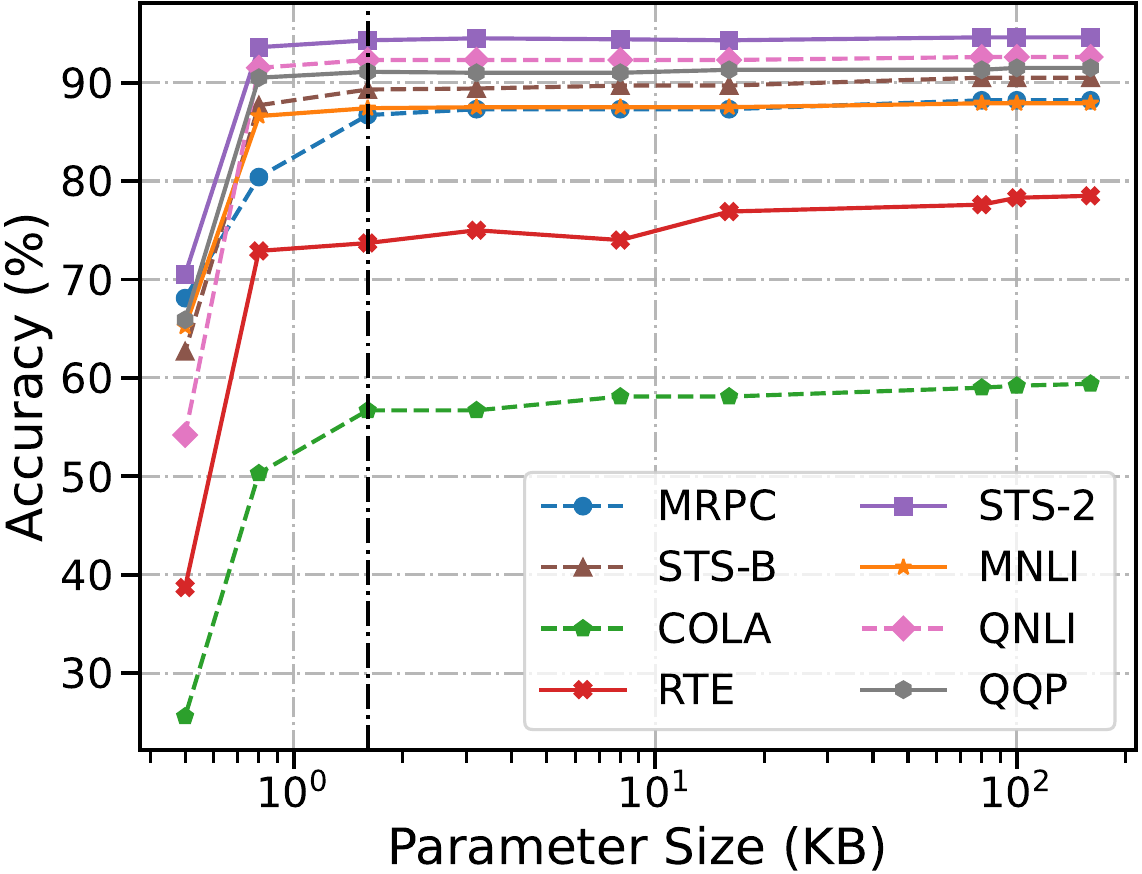}
            \caption{RoBERTa}
        \end{minipage}
        
    \caption{The accuracy of ResNet50 and RoBERTa on different datasets with different parameter size for elastic approximation.}
    \label{fig:extra_parameter}
\end{figure}

\subsubsection{CNN-based ResNet} For CNN-based ResNet, we compare the proposed \textit{NeuralMatrix} with ONESA \cite{sun2024onesa}, a method designed for computational efficiency, as well as with Cutout \cite{devries2017improved}, AAAI-21 \cite{zhao2021distribution}, and UI8 \cite{zhu2020towards}, which focus on improving accuracy at the cost of additional computation. These related works report results on datasets such as CIFAR-10, CIFAR-100, ImageNet, and Fashion-MNIST.

As illustrated in Fig. \ref{fig:accuracy_cnn}, \textit{NeuralMatrix} consistently outperforms ONESA in all cases. Compared to the baseline, \textit{NeuralMatrix} incurs a slight accuracy loss of 0.17\% to 0.39\%. However, when compared to Cutout, AAAI-21, and UI8, methods designed for higher inference accuracy rather than computational efficiency, the \textit{NeuralMatrix} can achieve better accuracy on occasional datasets.

\begin{table}[t]
\begin{center}
\begin{threeparttable}
\begin{tiny}
  \centering
  \caption{Parameter overhead comparison between \textit{NeuralMatrix} and relevant designs.}
  \setlength{\tabcolsep}{2.3pt}
    \begin{tabular}{ccccccc}
    \toprule
    \textbf{Designs} & NN-LUT\tnote{1} & ONESA\tnote{2}  & NPE\tnote{3}    & MA-BERT\tnote{4}  &\textbf{\textit{NeuralMatrix}} & \textbf{\textit{NeuralMatrix}} \\
    \midrule
    \textbf{DNN Model} & RoBERTa & BERT  & BERT & BERT & \textbf{ResNet-50} & \textbf{RoBERTa} \\
    \textbf{Precision} & FP32 & INT16  & INT8 & FP32 & \textbf{INT8} & \textbf{INT8} \\
    \midrule
    \textbf{Average Acc. Loss} &   0.28\%    &    0.88\%   &    2.55\%   &  0.13\% &  \textbf{0.12\%}    &  \textbf{0.40\%} \\
    \textbf{Ex. Parameter Size} & 100KB &   36KB    &   72KB    &   65.5KB  & \textbf{25KB}  &  \textbf{25KB} \\
    \bottomrule
    \end{tabular}%
  \begin{tablenotes}
  \item[1,2,3,4] The related design we compare our work with includes NN-LUT \cite{yu2022nn}, ONESA \cite{sun2024onesa}, NPE \cite{khan2021npe} and MA-BERT\cite{ming2022ma}
  \end{tablenotes}
  \label{tab:parameter_overhead}%
\end{tiny}
\end{threeparttable}
\end{center}
\end{table}%}

\subsubsection{Transformer-based BERT}
For the transformer-based BERT, which is more computationally intensive, we compare the proposed \textit{NeuralMatrix} with related works that focus on optimizing computational efficiency. In addition to ONESA \cite{sun2024onesa}, we consider MA-BERT \cite{ming2022ma}, which uses BERT, and NN-LUT \cite{yu2022nn} and I-BERT \cite{kim2021bert}, which use RoBERTa. To ensure a fair comparison, we implement both BERT and RoBERTa as baselines. Since MA-BERT does not incorporate quantization, we compare our quantization results with those of other related works. Additionally, we constrain the extra parameters introduced by the proposed \textit{NeuralMatrix} to be smaller than those in NN-LUT (100KB).

For FP32, \textit{NeuralMatrix} exhibits an average accuracy loss of 0.32\% compared to the RoBERTa baseline, while outperforming both NN-LUT and MA-BERT, which rely on neural network approximations, across five benchmarks. Moreover, compared to ONESA, which uses piecewise linear approximation, our approach improves average accuracy by 5.02\%.
For INT8, using I-BERT as the baseline, \textit{NeuralMatrix} incurs an average accuracy loss of just 0.11\%, while outperforming NN-LUT in four benchmarks and improving average accuracy by 0.85\% compared to ONESA.

In summary, for both CNN and transformer-based models, the proposed \textit{NeuralMatrix} exhibits a 0.11\% to 0.39\% reduction in accuracy compared to the baseline. When compared to related works that share the same efficiency design goals, \textit{NeuralMatrix} improves accuracy by 0.85\% to 5.02\%. Even when compared to methods that prioritize accuracy at the expense of additional computation costs, \textit{NeuralMatrix} can achieve better accuracy on occasional tasks.

\begin{figure}[t]
\centering
\includegraphics[width = 0.4\textwidth]{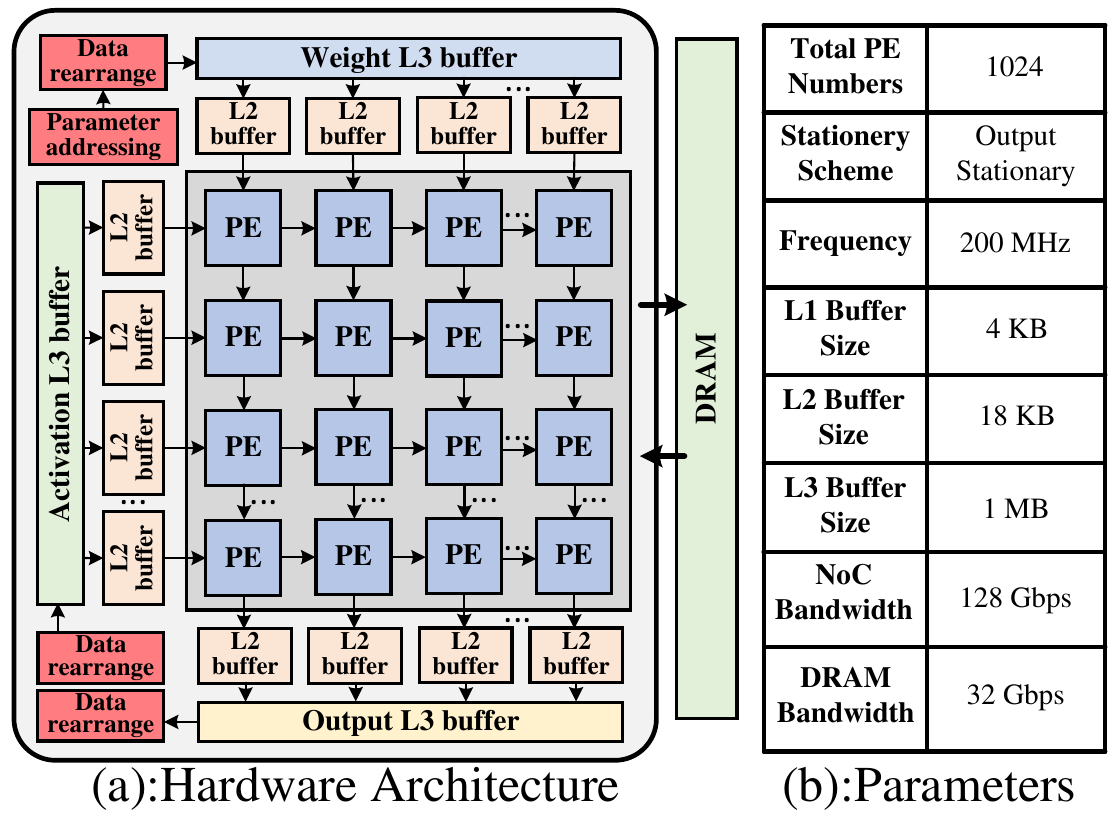}
\caption{Overview of the general matrix multiplication (GEMM) accelerator architecture.}
\label{fig:architecture}
\end{figure}

\begin{figure*}[t]
     \centering  
        \begin{minipage}[c]{0.45\textwidth}
            \captionsetup{type=subfigure}
            \includegraphics[width=\textwidth]{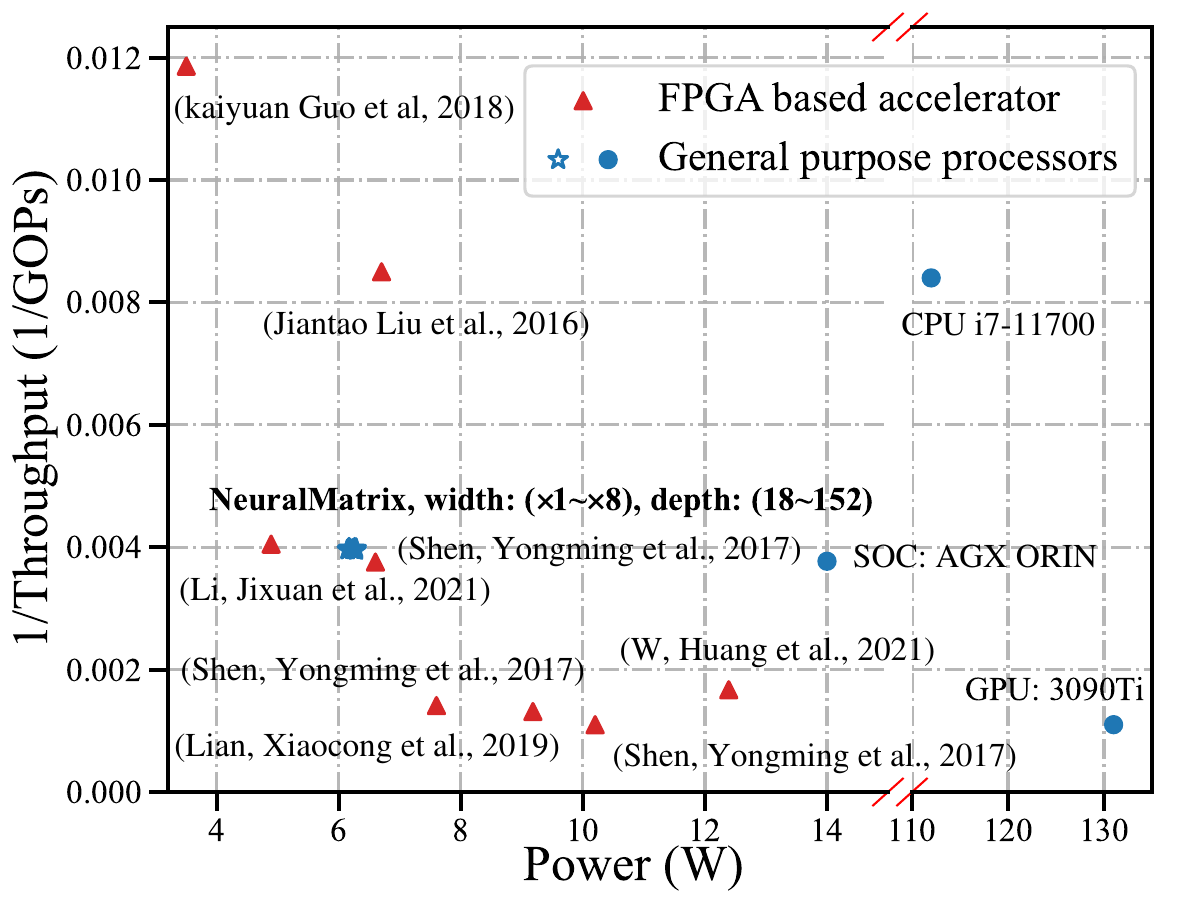}
            \caption{CNN (ResNet-50) and its variant models}
        \end{minipage}  
        \begin{minipage}[c]{0.45\textwidth}  
            \captionsetup{type=subfigure}
            \includegraphics[width=\textwidth]{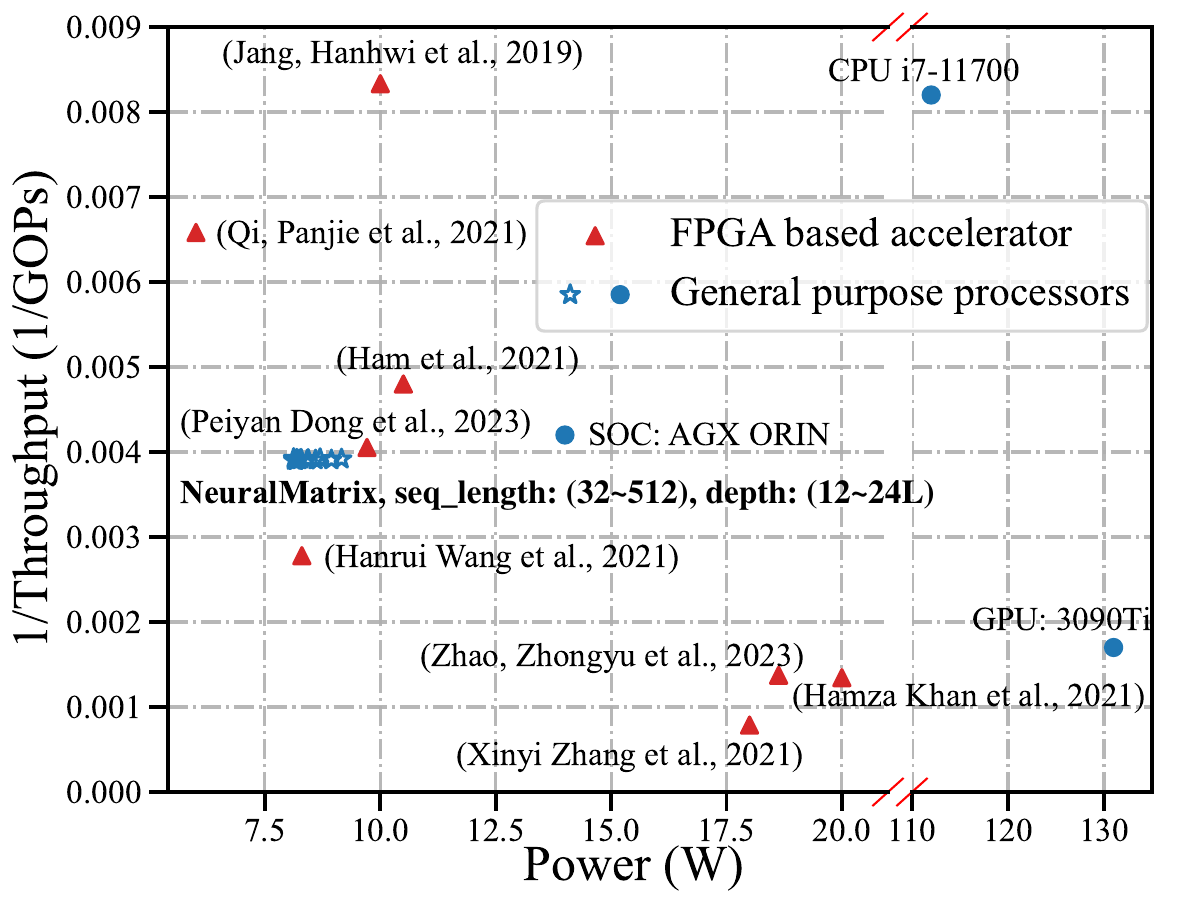}
            \caption{Transformer (BERT) and its variant models}
        \end{minipage}
        
    \caption{Different networks' computation efficiency (throughputs and power consumption) on different computing processors.}
    \label{fig:Computation_efficiency}
\end{figure*}

\subsection{Extra Model Parameter Sizes Comparison}
\label{sec:eval-parameter}
As the granularity of approximation introduces varying additional parameters, influencing the approximation accuracy and, consequently, the overall neural network performance, we conduct an evaluation of network inference accuracy across different approximation granularities, measured by the extra parameters introduced by the approximation, as depicted in Fig. \ref{fig:extra_parameter}. For both CNN-based ResNet and transformer-based BERT, the proposed \textit{NeuralMatrix} quickly achieves stable inference accuracy with 3.2KB, 1.6KB extra parameters for ResNet50 and RoBERTa.

% Table \ref{tab:parameter_overhead} illustrates the overheads introduced by the extra parameters from NeuralMatrix with the original network's parameter sizes. Notably, compared to the parameters in the original network, NeuralMatrix introduces only 0.01\% and 0.46\% additional parameters, demonstrating its practicality and efficiency in minimizing parameter overhead while maintaining stable network performance.

In Table \ref{tab:parameter_overhead}, we compare the additional parameter sizes of the proposed \textit{NeuralMatrix} with those of related works that also optimize the computation types in neural networks as described above. Here we use the baseline accuracy of RoBERTa mentioned in \cite{yu2022nn} to calculate the average accuracy loss for those designs based on RoBERTa model, and we use the baseline accuracy of BERT in this paper to calculate the accuracy loss for BERT designs. Given similar inference accuracy, the proposed \textit{NeuralMatrix} can reduce the additional parameter sizes by 1.44 to 4$\times$, benefiting from the elastic approximation with horizontal size optimizations and vertical bias correction. For example, in comparison to the extra parameters in NN-LUT, \textit{NeuralMatrix} reduces the number of extra parameters by approximately 4$\times$. Besides, the extra parameters account for only 0.06\% and 0.46\% of the parameters in the RoBERTa and ResNet50 models respectively, which illustrates its practicality in minimizing parameter overheads.

\subsection{Computation Efficiency with GEMM}
\label{sec:design}

%\begin{table}
%\begin{center}
%\begin{small}
%\caption{Parameter overhead in NeuralMatrix.}
%%\setlength{\belowcaptionskip}{-2mm}
%\label{tab:parameter_overhead}
%\begin{tabular}{ccc}
%\toprule  
%DNN Model & ResNet-50(Ours) & BERT(Ours) \\
%\midrule
%% Average Accuracy Loss (FP16) & 2.02\% & 1.32\% & 0.40\% \\ 
%% \hline
%% Average Accuracy Loss (INT8) & 2.97\% & 1.83\% & 1.80\% \\ 
%% \hline
%Extra Parameter Size (FP32) & 114KB & 49.2KB \\ 
%Extra Parameter Size (INT8) & 57KB & 24.6KB \\ 
%Normalized Parameter Size & 0.46\% & 0.01\%  \\ 
%\bottomrule
%\end{tabular}
%\end{small}
%\end{center}
%\end{table}

\textit{NeuralMatrix} enables the computation of entire neural networks using matrix operations, paving the way for executing complete neural networks on a General Matrix Multiplication (GEMM) accelerator. We conduct a comparative analysis of computational efficiency across various processing units, including general-purpose CPUs, GPUs, SoCs, FPGA-based Application Specific Integrated Circuits (ASICs), and the proposed \textit{NeuralMatrix} implemented with an FPGA-based GEMM accelerator.

\subsubsection{Implementation and Setup}
In this study, we implement a systolic array based GEMM accelerator with the Xilinx Virtex 7 XC7VX485T FPGA, as shown in Fig. \ref{fig:architecture}(a). The architectural-level parameters, including the number of Processing Elements (PEs) and memory bandwidths of the GEMM accelerator, are optimized with the GEMM design automation tool \citep{wei2017automated}. The parameters are summarized in Fig. \ref{fig:architecture}(b). We choose the output stationary data movement strategy as it avoids frequent data movements to and from memories and benefits the lowest energy consumption for large matrix multiplication \citep{zhao2022fpga}. 
To assess the computation performance of CPUs and GPUs, we conducted tests on the Intel i7-11700 CPU, NVIDIA 3090Ti GPU, and Jetson Orin SoC, utilizing an IT9121 power analyzer. For the existing FPGA-based ASICs, we gathered relevant data from published papers. To standardly and normally compare the computation performance across different network models and hardware processors. Our analysis mainly focuses on the computation efficiency, which is indicated by the computation throughput (i.e., operations per second) with the power consumption from the hardware processor.
Higher throughput with smaller power consumption indicates better computation efficiency.

\subsubsection{Computation Efficiency}

Fig. \ref{fig:Computation_efficiency} illustrates the computation efficiency (recorded with 1/throughput and power) of CNN-based ResNet-50, transformer-based BERT and their variants on different hardware processors. Each point indicates a network on a hardware processor. 
All the design points are scatter-plotted and the processor types are distinguished by the marker shapes.

Clearly, across all the network models, the general-purpose processors CPU and GPU, especially the CPU, are located far away from the Pareto frontiers of design points, indicating a low computation efficiency. This is because the general-purpose processors trade the computation efficiency for generality by incorporating too many function units. To quantify the improvement with CPU, GPUs, SoC, the \textit{NeuralMatrix} improves computation efficiency (i.e., throughput per power) by 38.72, 5.93, and 2.17$\times$ for the CNN-based ResNet-50 and by 28.96, 7.02, and 1.85$\times$ for the transformer-based RoBERTa.

% 2019DNNBuilder,2021CARLA,2019Fast,2020Low,2016Eyeriss,2020Look,2020A,
%

The plots also indicate the related FPGA-based ASIC designs for these networks in recent years \citep{2019High, guo2017angel, 8192499, 8980322, 9848824, 9634838, 9643586, khan2021npe, wang2021spatten, dong2023heatvit, 10.1145/3477002, 2020A, 10.1145/2847263.2847265}. Compared to ASIC designs for specific networks, the \textit{NeuralMatrix} can achieve the same level of computation efficiency distributed along the Pareto frontiers. Please note that the proposed \textit{NeuralMatrix} can execute neural networks efficiently and benefit the generality of different neural networks.

When comparing \textit{NeuralMatrix}'s computation efficiency across network variants, we find that efficiency increases with wider and deeper networks, with width having a greater impact than depth. Efficiency peaks and stabilizes for larger networks. This finding shows that \textit{NeuralMatrix} is especially effective for large, wide networks, as the larger matrices enable greater parallelism in the GEMM accelerator.

% \begin{table}[t]
%  \begin{center}
%  \begin{tiny}
% % \begin{footnotesize}
% \caption{{The parameters of the implemented GEMM accelerator.}}
% % \setlength{\belowcaptionskip}{-2mm}
% \label{tab:GEMM_config}
% %\resizebox{\columnwidth}{!}{
% \begin{tabular}{cccc}
% \toprule  
%  \makecell{\textbf{Total PE Numbers}} & \makecell{\textbf{Cluster Numbers} } & \makecell{\textbf{Stationary Scheme} } & \makecell{\textbf{Frequency}}\\
% % \hline
% 1024 & 8 & Output stationary & 200 MHz \\
% \midrule
% \makecell{\textbf{L1 buffer Size} 
% } & \makecell{\textbf{L2 buffer Size} } & \makecell{\textbf{NoC Bandwidth} } & \makecell{\textbf{DRAM Bandwidth} }\\
% % \hline
% 4 KB & 1 MB & 128 Gbps & 32 Gbps \\
% \bottomrule
% \end{tabular}
% %}
% % \end{footnotesize}
%  \end{tiny}
%  \end{center}
% \end{table}

\section{Conclusion}
\label{sec:conclusion}
To address the challenge posed by the different types of computations and facilitate the deployment of versatile deep neural networks all on a General
Matrix Multiplication (GEMM) accelerator, we introduce NeuralMatrix. 
This NeuralMatrix transforms entire DNN architectures into linear matrix operations, which can then be efficiently executed by GEMM accelerator. The mainstream DNN backbone models CNN and transformer, along with their variant models, are evaluated as illustrative examples. Experiments reveal that transitioning the entire neural network to linear matrix operations results in negligible inference accuracy loss and demonstrates application-specific levels of computation and energy efficiency when paired with GEMM accelerators. 

\bibliography{aaai25}
\clearpage
\end{document}